\documentclass[letterpaper, 10 pt, conference]{ieeeconf}  

\IEEEoverridecommandlockouts                              

\usepackage{hyperref}
\hypersetup{
    colorlinks=true,   
    urlcolor=blue,     
    linkcolor=black,   
    citecolor=black    
}
\title{\LARGE \bf
TouchBridge: Tactile-Conditioned Human-to-Robot Transfer for Dexterous Manipulation}

\author{%
Haozhe Zhang\textsuperscript{1*},
Gu Zhang\textsuperscript{1*†},
Zirui Zou\textsuperscript{2*},
Yuzheng Chen\textsuperscript{3},
Jinlong Wu\textsuperscript{4},\\
Zixuan Hu\textsuperscript{5},
Jianke Zhang\textsuperscript{1},
Yucheng Hu\textsuperscript{1},
Zhecheng Yuan\textsuperscript{1},
Huazhe Xu\textsuperscript{1}\\[1ex]
{\small
\textsuperscript{1} Tsinghua University}
{\small
\textsuperscript{2} Zhejiang University}
{\small
\textsuperscript{3} Nanyang Technological University}
{\small
\textsuperscript{4} Hunan University}
{\small
\textsuperscript{5} Peking University}\\
{\small
\textsuperscript{*} Equal Contribution}\space
{\small
\textsuperscript{†} Project Lead}\\
{\small
\textbf{Project Website}: \url{https://touchbridge2026.github.io/}
}
}

\usepackage{url} 
\usepackage{booktabs}
\usepackage{amsmath}
\usepackage{graphicx}
\usepackage{float}
\usepackage{capt-of}
\usepackage{cite}
\usepackage{pifont}
\usepackage{makecell}
\usepackage{tabularx}
\usepackage{array}
\newcommand{\cmark}{\ding{51}}
\newcommand{\xmark}{\ding{55}}
\DeclareRobustCommand{\xovercmark}{%
  \mbox{%
    \ooalign{%
      \hidewidth\raisebox{0.1ex}{\xmark}\hidewidth\cr
      \cmark\cr
    }%
  }%
}

\usepackage{caption}

\begin{document}

\IEEEaftertitletext{%
    \begin{minipage}{\textwidth}
        \centering
        \includegraphics[width=0.93\linewidth]{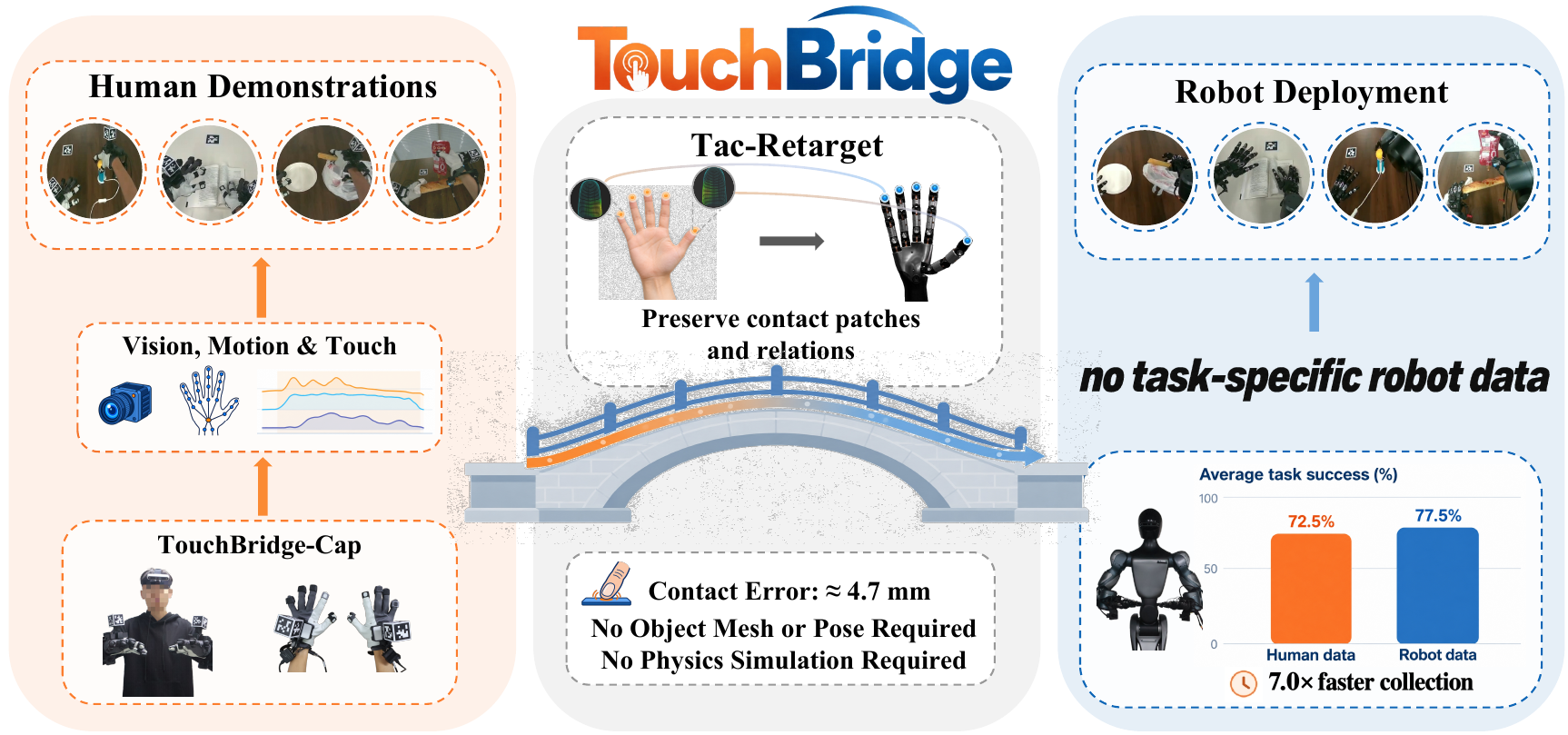}
        \captionof{figure}{We introduce TouchBridge, a tactile-conditioned human-to-robot transformation framework. TouchBridge-Cap
records human demonstrations with
sensing configurations aligned to the robot interface.
Building on this correspondence, Tac-Retarget converts
human demonstrations into robot-domain training data while
preserving the demonstrated contact locations and spatial
relations. Policies fine-tuned on these data achieve success
rates close to those achieved by policies trained on teleoperated robot
demonstrations, while human demonstrations can be collected
substantially faster.}
        \label{fig:teaser}
    \end{minipage}
}

\maketitle
\thispagestyle{empty}
\pagestyle{empty}


\begin{abstract}
Human demonstrations offer an efficient and scalable source of data for dexterous robot manipulation, but differences in sensing and hand morphology complicate the transfer of contact-rich interactions. We present \textbf{TouchBridge}, a hardware--software co-designed framework for human-to-robot data transformation. A wearable human demonstration interface and a corresponding robot interface establish cross-embodiment correspondence in visual and fingertip tactile sensing. We introduce \textbf{Tac-Retarget}, a tactile-conditioned retargeting method that uses measured contact to align fingertip contact patches and preserve their spatial relations within and across hands. Unlike object-aware or simulation-based transfer pipelines, Tac-Retarget requires no object mesh, object-pose estimate, or physics simulation for retargeting, enabling direct and scalable conversion of human demonstrations into robot-domain training data. Combined with visual synthesis and camera-anchored action conversion, TouchBridge produces robot-domain demonstrations for fine-tuning a pretrained vision-language-action (VLA) model. Across four real-world contact-rich tasks, policies fine-tuned solely
on TouchBridge-converted human demonstrations achieve a 72.5\% average success rate, compared with 77.5\% using an equal number of teleoperated robot demonstrations, while human demonstration collection is 7.0$\times$ faster. Under the same policy and training setup, Tac-Retarget outperforms DexPilot and UniDex by 25.0 and 42.5 percentage points, respectively.

\end{abstract}

\section{INTRODUCTION}

Building foundation models for contact-rich dexterous manipulation requires large-scale demonstrations that capture not only hand motion, but also physical interaction with the environment. Collecting such data directly on dexterous robots, however, is slow and demands substantial operator effort. Human demonstrations provide a natural and potentially scalable source of manipulation experience, and have increasingly been used for large-scale pretraining~\cite{zhang2026unidex,zheng2026egoscale} and direct human-to-robot transfer~\cite{wang2026humanego,guzey2026dexterity}. Realizing this potential, however, requires transforming human experience into robot-executable training data both faithfully and efficiently.

This transformation faces two key challenges. First, \emph{kinematic correspondence alone does not preserve physical interaction}. Conventional retargeting methods typically match hand poses, fingertip positions, or hand-centric vectors~\cite{handa2020dexpilot,qin2023anyteleop,zhang2026unidex}. While effective for free-space motion, these objectives do not explicitly capture contact. As a result, fixed human-derived fingertip targets can displace a contacting finger from its support region, distort relations among simultaneous contacts, or break contact altogether. For contact-rich manipulation, kinematically plausible motion may therefore still be physically ineffective.

Second, \emph{preserving physical interaction should not come at the cost of data scalability}. Object-aware approaches incorporate hand--object geometry or physics-based constraints~\cite{huang2025fungrasp,lin2025dexflow,wu2026toporetarget,fan2026robopaint}, while other human-to-robot pipelines refine transferred trajectories or policies through simulation and reinforcement learning~\cite{li2025maniptrans, kedia2026simtoolreal}. Such methods can improve physical consistency, but require object meshes, object-pose estimation, environment reconstruction, or simulation-based rollout and refinement. These object- and task-specific requirements introduce substantial modeling and engineering overhead, making them difficult to scale with increasingly large and diverse human datasets.

Our key observation is that \emph{physical contact itself provides a natural correspondence across embodiments}. Tactile sensing directly reveals which fingertips are in contact, where contact occurs, and how strongly different regions are loaded. Because these signals are measured directly during human demonstrations, they provide interaction constraints without requiring explicit object reconstruction or simulation-based refinement. Tactile sensing therefore addresses both challenges simultaneously: it augments kinematic retargeting with physical interaction information while preserving a direct human-to-robot data conversion pathway.

Building on this idea, we present \textbf{TouchBridge}, a hardware–software co-designed framework for scalable conversion of human demonstrations of contact-rich dexterous manipulation into robot-domain data. At the hardware level, we develop \textbf{TouchBridge-Cap}, a wearable capture interface that records RGB observations, hand motion, and fingertip tactile measurements. The corresponding robot interface approximately matches the wrist-camera-to-hand geometry, fingertip sensor placement, and tactile sensing layout of TouchBridge-Cap, establishing cross-embodiment correspondence in visual and tactile observations. At the software level, TouchBridge transforms these human demonstrations into robot-domain observations and actions for policy learning without task-specific robot demonstrations or object-specific environment modeling.

At the core of TouchBridge is \textbf{Tac-Retarget}, a tactile-conditioned retargeting method that uses measured contact to guide motion transfer across embodiments. Tac-Retarget combines \emph{contact-patch alignment}, which aligns tactile contact regions between human and robot fingertips, with a \emph{contact-activated graph}, which preserves spatial relations among simultaneous contacts within and across hands. Together, these objectives transfer not only fingertip motion but also the geometry and spatial structure of demonstrated contacts. The optimization operates directly on measured human kinematics and tactile responses, requiring no object mesh, object-pose estimate, or physics simulation. Combined with robot-domain visual synthesis and camera-anchored action construction, the resulting pipeline converts raw human demonstrations into robot-domain training sequences.

We evaluate TouchBridge on four contact-rich real-world manipulation tasks: bulb screwing, sauce application, fragile-object grasping, and page flipping. Under the same policy and training setup, demonstrations converted using Tac-Retarget yield an average real-robot policy success rate of $\mathbf{72.5\%}$, outperforming DexPilot~\cite{handa2020dexpilot} and UniDex~\cite{zhang2026unidex} by $25.0$ and $42.5$ percentage points, respectively, while also achieving lower contact error. Policies fine-tuned solely on TouchBridge-converted human demonstrations achieve a $72.5\%$ average success rate, close to the $77.5\%$ obtained using an equal number of teleoperated robot demonstrations, while human demonstrations are collected $\mathbf{7.0}\times$ faster. These results show that TouchBridge retains much of the utility of robot-collected demonstrations while substantially reducing the cost of acquiring training data, providing a practical path toward scaling robot learning from human experience.

Our contributions are summarized as follows:
\begin{itemize}
\item We present \textbf{TouchBridge}, a hardware--software co-designed framework that transforms contact-rich human demonstrations into robot-domain training sequences. Through aligned visual and fingertip tactile sensing across human and robot interfaces, TouchBridge enables policy learning for contact-rich dexterous manipulation without task-specific teleoperated robot demonstrations.

\item We introduce \textbf{Tac-Retarget}, a tactile-conditioned retargeting method that uses measured contact-patch alignment and a contact-activated graph to preserve physical interaction across embodiments. Compared with kinematic retargeting, Tac-Retarget substantially improves both contact preservation and downstream real-robot policy success.

\item TouchBridge converts human demonstrations without object meshes, object-pose estimates, or simulation-based rollout refinement. Together, the $\mathbf{7.0}\times$ faster demonstration collection and policy performance close to that obtained from robot-collected data make the framework well suited for scaling robot learning from human demonstrations.

\end{itemize}

\section{Related Work}

\subsection{Learning from Human Demonstrations}
Human demonstrations provide a scalable alternative to collecting task
data on physical robots. UniDex~\cite{zhang2026unidex} and
EgoScale~\cite{zheng2026egoscale} use large-scale human-derived data
to pretrain transferable dexterous policies. These works primarily emphasize scaling human-derived data,
with less emphasis on the fidelity of the converted
demonstrations, particularly the preservation of contact
interactions across embodiments. EgoMimic~\cite{kareer2025egomimic} and
MotionTrans~\cite{yuan2026motiontrans} co-train
manipulation policies on human and robot demonstrations.
Many other methods use simulation to ground human demonstrations in the robot
embodiment, building on the success of sim-to-real reinforcement
learning~\cite{wang2025boostergym, wang2026soccer}. ManipTrans~\cite{li2025maniptrans} refines retargeted
hand-object trajectories through residual reinforcement learning.
SimToolReal~\cite{kedia2026simtoolreal} and Dex-X~\cite{chen2026dexx} train policies
in simulation and deploy them zero-shot, using human videos as references for object
motion or hand--object interaction; Dex-X further relies on simulated
contact~\cite{si2024difftactile} to supply the tactile signals absent from human videos. These methods involve additional effort in environment construction
and robot--object interaction modeling. A further line of work develops interfaces and transformation
pipelines for learning directly from human demonstrations. DexUMI~\cite{xu2025dexumi} combines wearable exoskeletons and visual conversion to bridge the human--robot embodiment gap, with the exoskeleton mechanically designed to match the target hand's kinematics. Adapting to new robot hands generally requires hardware
redesign, which is particularly challenging for high-DoF hands. Moreover, the wearable mechanism is typically tailored to a limited range of human hand sizes, reducing user adaptability and potentially compromising wearer comfort during demonstration collection.
HumanEgo~\cite{wang2026humanego} enables task-level transfer
without robot training data, but has only been demonstrated
with parallel-jaw grippers. AINA~\cite{guzey2026dexterity} extends robot-data-free
transfer to multi-fingered hands using visually estimated
hand--object geometry and fingertip motion, without
explicit tactile contact constraints.
Our method instead uses measured tactile signals to preserve
contact patches and their spatial relations during
motion retargeting without simulation refinement, while using a glove-based capture
interface that requires no mechanical correspondence
with the target hand's joint structure. This enables policies fine-tuned
solely on converted human demonstrations to be deployed on
the real robot without task-specific robot demonstrations.

\begin{table}[t]
    \centering
    \footnotesize
    \setlength{\tabcolsep}{1.5pt}
\caption{\textbf{Comparison of human-to-robot transfer frameworks.} High-DoF denotes hands with at least 20 DoF. Robot-Free denotes fine-tuning without robot-collected demonstrations. \xovercmark~indicates that the capability
has not been experimentally validated.}
    
    \renewcommand{\arraystretch}{1.15}

    \begin{tabularx}{\columnwidth}{
        @{}
        >{\raggedright\arraybackslash}X
        >{\centering\arraybackslash}p{0.14\columnwidth}
        >{\centering\arraybackslash}p{0.13\columnwidth}
        >{\centering\arraybackslash}p{0.15\columnwidth}
        !{\hspace{4pt}}
        >{\centering\arraybackslash}p{0.17\columnwidth}
        @{}
    }
        \hline
        \noalign{\vskip 2pt}
        \textbf{Method} &
        \shortstack{\textbf{High-DoF}\\\textbf{Hands}} &
        \shortstack{\textbf{Sim.-}\\\textbf{Free}} &
        \shortstack{\textbf{No Object}\\\textbf{Pose/Geom.}} &
        \shortstack{\textbf{Robot-Free}\\\textbf{Fine-Tuning}} \\
        \hline

        \textbf{TouchBridge}
        & \cmark & \cmark & \cmark & \cmark \\

        AINA~\cite{guzey2026dexterity}
        & \xmark & \cmark & \xmark & \cmark \\

        HumanEgo~\cite{wang2026humanego}
        & \xmark & \cmark & \xmark & \cmark \\

        DexUMI~\cite{xu2025dexumi}
        & \xmark & \cmark & \cmark & \cmark \\

        ManipTrans~\cite{li2025maniptrans}
        & \cmark & \xmark & \xmark & \cmark \\

        TopoRetarget~\cite{wu2026toporetarget}
        & \cmark & \xmark & \xmark & \cmark \\

        RoboPaint~\cite{fan2026robopaint}
        & \xovercmark & \xmark & \xmark & \cmark \\

        SPIDER~\cite{spider}
        & \cmark & \xmark & \xmark & \cmark \\

        C2Dex~\cite{ren2026c2dex}
        & \xovercmark & \xmark & \xmark & \xovercmark \\

        MotionTrans~\cite{yuan2026motiontrans}
        & \xovercmark & \cmark & \cmark & \xmark \\

        EgoMimic~\cite{kareer2025egomimic}
        & \xmark & \cmark & \cmark & \xmark \\

        \hline
    \end{tabularx}
    
    \label{tab:method-comparison}
    \vspace{-15pt}
\end{table}

\subsection{Dexterous Retargeting}

Dexterous retargeting maps human hand motion to robot hands with
different kinematic structures~\cite{huang2026icl}. DexPilot~\cite{handa2020dexpilot}
and AnyTeleop~\cite{qin2023anyteleop} optimize hand-centric
task-space or keypoint-vector correspondences. UniDex~\cite{zhang2026unidex} combines fingertip inverse kinematics
with manual adjustment of the hand-base offset. These approaches are effective for pose transfer but do not condition the mapping
on tactile measurements, which have proven valuable for contact-rich
control~\cite{jiang2024cimpc} and policy learning~\cite{wang2026univlat}. FunGrasp~\cite{huang2025fungrasp} transfers
object-relative functional contacts and uses simulation-trained control
for execution. DexFlow~\cite{lin2025dexflow} refines retargeted poses
using contact maps inferred from hand-object geometry.
TopoRetarget~\cite{wu2026toporetarget} preserves local interaction
topology using an object mesh and pose trajectory.
C2Dex~\cite{ren2026c2dex} reconstructs stable object-side contacts from
monocular video and further refines the trajectory in simulation. SPIDER~\cite{spider} uses physics-based sampling and virtual
contact guidance to generate dynamically feasible robot
trajectories from human demonstrations. ... RoboPaint~\cite{fan2026robopaint} uses force-weighted tactile correspondences
within a Real-Sim-Real pipeline, and concurrent work CARD~\cite{dai2026card}
anchors object-free retargeting on glove-measured contact to obtain contact-force
targets for open-loop hardware replay.
Our method instead preserves measured contact patches together with their spatial
relations within and across hands, and converts human demonstrations into complete
robot-domain training data for policy learning, without relying on physical
simulation, object meshes, or object-pose estimates.

As shown in Tab.~\ref{tab:method-comparison}, compared with prior human-to-robot transfer frameworks, TouchBridge supports high-DoF dexterous hands ($20$ DoF) with simulation-free and object-model-free
data conversion, enabling task learning directly from converted human
demonstrations.

\begin{figure}[t]
    \centering
    \includegraphics[width=0.98\linewidth]
    {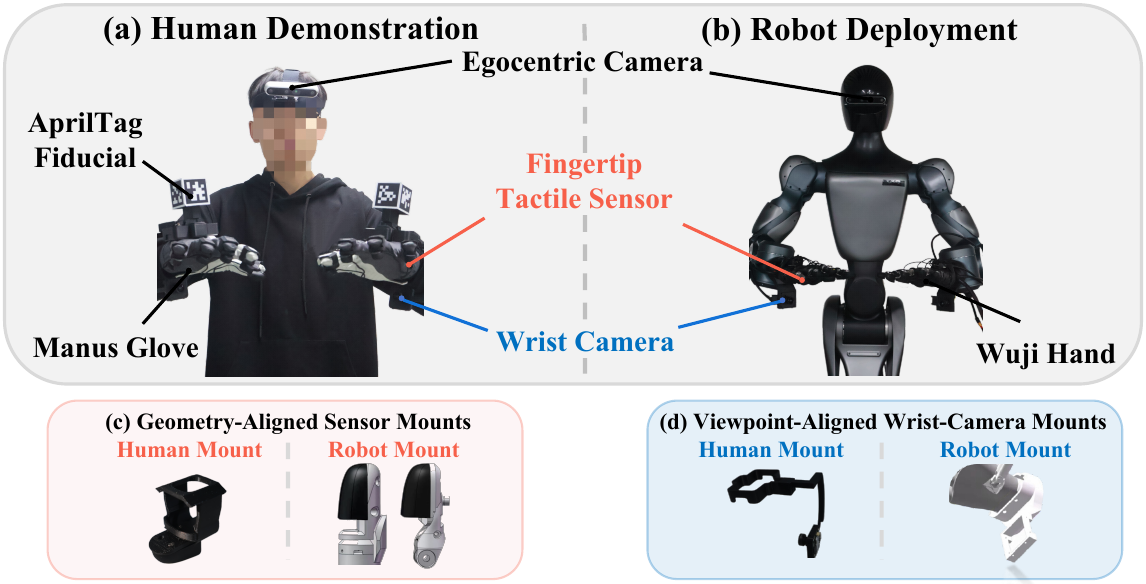}
    \caption{
        \textbf{Human--Robot Hardware Co-design.}
        (a,b) Human demonstration and robot deployment interfaces.
        (c,d) Corresponding fingertip
        tactile-sensor placements and wrist-camera configurations reduce mismatch across embodiments.
    }
    \label{fig:cross_embodiment_system}
    \vspace{-20pt}
\end{figure}

\section{Method}
\begin{figure*}[!t]
    \centering
    \includegraphics[width=0.95\textwidth]
    {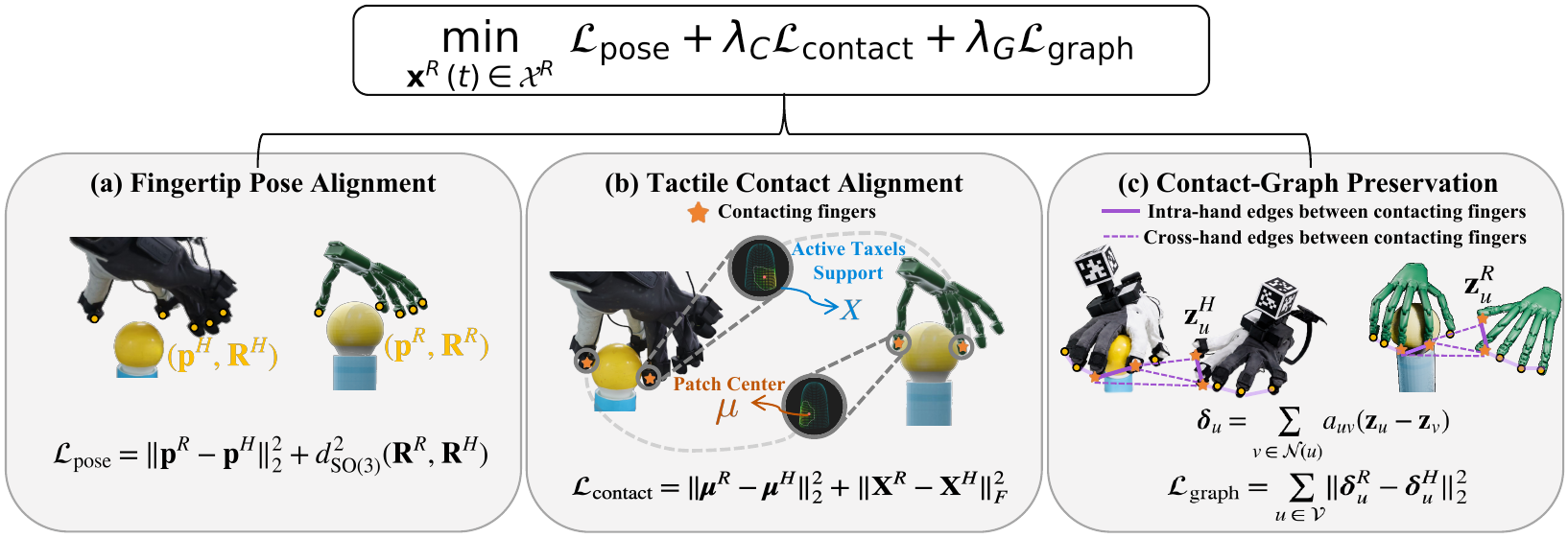}
    \caption{
        \textbf{Tac-Retarget}.
        (a) Fingertip pose alignment tracks the demonstrated
        fingertip positions and orientations in the camera frame.
        (b) Contact alignment matches the patch center and active
        taxels between the human and robot fingertips.
        (c) A tactile-activated contact graph preserves the
        demonstrated spatial relations among active contacts
        within and across the two hands.
        Weighting terms and regularization are omitted for clarity.
    }
    \label{fig:tactile_retarget}
    \vspace{-15pt}
\end{figure*}
\subsection{Human-Robot Vision and Tactile System}
\label{sec:cross_embodiment_visuotactile_system}

We align the wrist-camera and fingertip tactile-sensor configurations across the human demonstration and robot deployment interfaces. As shown in Fig.~\ref{fig:cross_embodiment_system}, both
interfaces use similarly mounted wrist cameras. Their
fingertip tactile sensors have corresponding placements
and matched sensing layouts. Differences in hand morphology and
egocentric appearance are handled by the human-to-robot transformation
pipeline described in Sec.~\ref{sec:human_to_robot}.

\subsubsection{Human Demonstration Interface}
\label{sec:human_interface}

We develop \textbf{TouchBridge-Cap}, a wearable human demonstration
interface that records egocentric RGB observations using an Intel
RealSense D455, together with bilateral wrist-view RGB observations
from two Hikrobot cameras. A pair of Manus gloves captures articulated
hand kinematics, while a PaXini Hall-effect tactile array mounted on each
fingertip records spatially distributed contact responses.
Multi-face AprilTag fiducials mounted near the wrists are localized
offline in the egocentric video. Using the detected fiducial poses and calibrated fiducial-to-glove transformations, we transform the glove-estimated fingertip poses into the D455 camera frame.

\subsubsection{Robot Deployment Interface}
\label{sec:robot_interface}

The robot is equipped with an egocentric camera, bilateral wrist cameras,
and five PaXini tactile sensors on each dexterous hand. Custom mounts position the wrist cameras similarly to those on \textbf{TouchBridge-Cap} and place tactile sensors at
corresponding fingertip locations. Together with the matched tactile sensing layouts,
this design provides aligned wrist-view observations and fingertip-local
tactile representations across the human and robot embodiments.

\subsection{Tac-Retarget}
\label{sec:tactile_retargeting}

\subsubsection{Problem Setup}
At time $t$, the human input consists of bimanual camera-frame fingertip poses $\mathbf{P}^{H}(t)=\big(\mathbf{p}^{H}(t),\mathbf{R}^{H}(t)\big)$ and synchronized distributed tactile measurements $\mathbf{S}^{H}(t)$, where $\mathbf{p}^{H}(t)$ and $\mathbf{R}^{H}(t)$ collect the fingertip positions and orientations, respectively. The robot state is $\mathbf{x}^{R}(t)=\big(\mathbf{T}^{R}(t),\mathbf{q}^{R}(t)\big)$, where $\mathbf{T}^{R}(t)\in\mathrm{SE}(3)^2$ contains the two hand-base poses and $\mathbf{q}^{R}(t)$ stacks the joint configurations of both hands. Forward kinematics maps this state to the robot fingertip poses
$\mathbf{P}^{R}(t)=\big(\mathbf{p}^{R}(t), \mathbf{R}^{R}(t)\big)$.

We solve
\begin{equation}
    \mathbf{x}^{R*}(t)
    =
    \arg\min_{\mathbf{x}^{R}(t)\in\mathcal{X}^{R}}
    \mathcal{L}_{\mathrm{pose}}
    +\lambda_{C}\mathcal{L}_{\mathrm{contact}}
    +\lambda_{G}\mathcal{L}_{\mathrm{graph}}
    +\mathcal{L}_{\mathrm{reg}},
    \label{eq:tactile_retarget}
\end{equation}
where $\mathcal{X}^{R}$ is the feasible bimanual robot state space. The first three terms correspond to the fingertip pose alignment, tactile contact alignment, and contact graph shown in Fig.~\ref{fig:tactile_retarget}, while $\mathcal{L}_{\mathrm{reg}}$ combines hand-base regularization, feasibility penalties and temporal regularization terms. No object mesh or pose estimate enters the optimization.

\subsubsection{Fingertip Pose Alignment}
We align the robot fingertip poses with the demonstrated human
fingertip poses in the camera frame. We use the calibrated tactile-sensor frames to define
fingertip poses for both motion tracking and contact alignment.
The pose objective is
\begin{equation}
\mathcal{L}_{\mathrm{pose}}
=
\sum_i w_i(t)
\left[
\|\mathbf{p}_i^R-\mathbf{p}_i^H\|_2^2
+
\lambda_{\mathrm{rot}}
d_{\mathrm{SO}(3)}^2(\mathbf{R}_i^R,\mathbf{R}_i^H)
\right],
\label{eq:fingertip_pose_alignment}
\end{equation}
where $i$ indexes the fingertips of both hands, $w_i(t)$ denotes
the contact-conditioned tracking weight, and
$d_{\mathrm{SO}(3)}$ is the geodesic distance between rotations.
The position and orientation terms establish global motion
correspondence across embodiments. Both robot hand-base poses
remain optimization variables rather than tracking targets
derived from the measured human wrists.

\subsubsection{Tactile Contact Alignment}
For each fingertip, contact is detected from its total
tactile response and number of active taxels. For a contacting fingertip $i$, let $\mathbf{r}_{im}$ denote
the calibrated local position of taxel $m$ and $s_{im}(t)$
its tactile response magnitude. The response-weighted
contact center in the fingertip sensor frame is
\begin{equation}
    \boldsymbol{\mu}_{i,\mathrm{loc}}(t)
    =
    \frac{
        \sum_m s_{im}(t)\mathbf{r}_{im}
    }{
        \sum_m s_{im}(t)+\epsilon
    }.
    \label{eq:patch_center}
\end{equation}

With matched calibrated taxel layouts, we use the same
human-derived local contact center for both human and robot
fingertips. We transform this center into the camera frame
using the measured human sensor pose and the robot sensor
pose from forward kinematics, obtaining
$\boldsymbol{\mu}_i^H$ and $\boldsymbol{\mu}_i^R$, respectively. We stack these camera-frame centers over contacting
fingertips as $\boldsymbol{\mu}^H(t)$ and
$\boldsymbol{\mu}^R(t)$. Similarly, $\mathbf{X}^H(t)$ and
$\mathbf{X}^R(t)$ collect corresponding active-taxel
positions transformed into the camera frame. The contact objective is
\begin{equation}
\mathcal{L}_{\mathrm{contact}}
=
\left\|
\boldsymbol{\mu}^{R}(t)-\boldsymbol{\mu}^{H}(t)
\right\|_2^2
+
\lambda_x
\left\|
\mathbf{X}^{R}(t)-\mathbf{X}^{H}(t)
\right\|_{\boldsymbol{\Omega}(t)}^2,
\label{eq:tactile_contact_alignment}
\end{equation}
where $\boldsymbol{\Omega}(t)$ contains normalized
active-taxel response weights, assigning greater importance
to taxels with stronger responses. The first term aligns the response-weighted patch centers, while the second preserves the spatial distribution of the active taxels.

\subsubsection{Contact Graph}
To preserve the spatial relations among simultaneous contacts, we construct corresponding human and robot contact graphs. We follow the distance-weighted Laplacian matching formulation
of TopoRetarget~\cite{wu2026toporetarget} and adapt it to a
tactile-activated graph. Within each hand, intra-hand edges connect pairs of contacting
fingertips, while cross-hand edges connect active contacts between the two hands. Tactile activation determines this contact-dependent topology, which is shared by the human and robot graphs.

Let $\mathbf{z}^{H}_u$ and $\mathbf{z}^{R}_u$ denote the corresponding human and robot positions of node $u$. The edge weights are computed and normalized as
\begin{equation}
\begin{gathered}
\tilde{a}_{uv}
=
\exp\left(
-\kappa
\left\|
\mathbf{z}^{H}_u-\mathbf{z}^{H}_v
\right\|_2
\right)
\left[
1+\gamma
\left(
\phi_u+\phi_v
\right)
\right],
\\[4pt]
a_{uv}
=
\frac{\tilde{a}_{uv}}
{\sum_{k\in\mathcal{N}(u)}\tilde{a}_{uk}+\epsilon},
\end{gathered}
\label{eq:tactile_graph_weight}
\end{equation}
where $\phi_u$ is the tactile response associated with node $u$. The graph topology and normalized weights are computed from the human demonstration and applied unchanged to the robot graph. For each node, we summarize its local spatial relations by the weighted relative-offset vector
\begin{equation}
    \boldsymbol{\delta}_u
    =
    \sum_{v\in\mathcal{N}(u)}
    a_{uv}
    \left(
    \mathbf{z}_u-\mathbf{z}_v
    \right), 
    \mathcal{L}_{\mathrm{graph}}
    =
    \sum_u
    \left\|
    \boldsymbol{\delta}^{R}_u
    -
    \boldsymbol{\delta}^{H}_u
    \right\|_2^2.
    \label{eq:tactile_graph}
\end{equation}
Contact determines which interaction edges are active, while tactile magnitude controls their relative importance. Matching the weighted relative offsets preserves the demonstrated spatial relations among active contacts within and across the two hands.

\subsection{Human-to-Robot Data Transformation Pipeline}
\label{sec:human_to_robot}

\begin{figure}[!t]
    \centering
    \includegraphics[width=\linewidth]{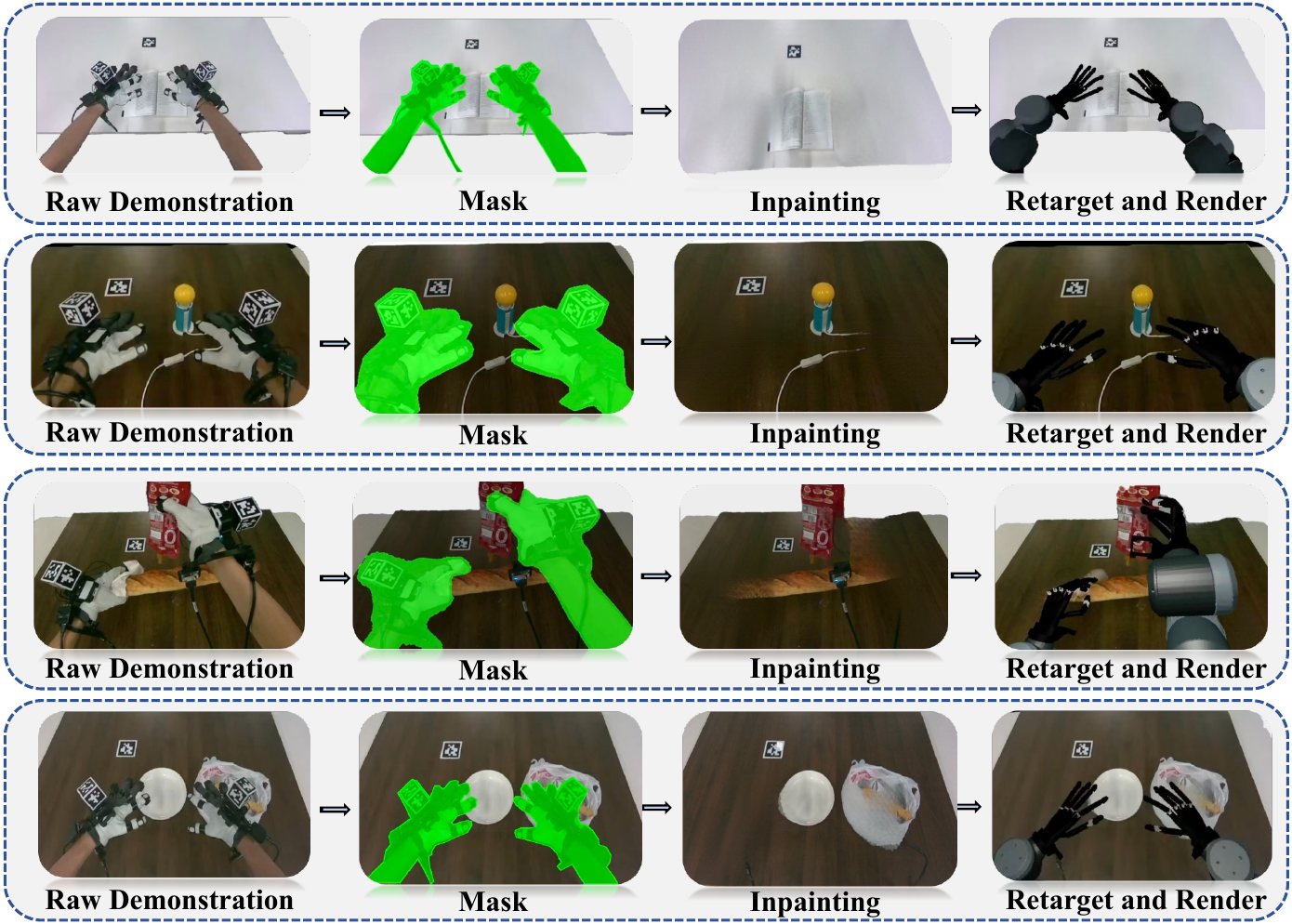}
    \caption{\textbf{Human-to-robot data generation.}
From left to right: raw human demonstrations, removal masks, inpainted scenes, and synthesized robot-domain observations.}
    \label{fig:human_to_robot}
    \vspace{-15pt}
\end{figure}

\subsubsection{Robot-Domain Observation Synthesis}
\label{sec:robot_domain_synthesis}

As shown in Fig.~\ref{fig:human_to_robot}, we synthesize
robot-domain visual observations from human demonstrations.
Building on the hardware design described in
Sec.~\ref{sec:cross_embodiment_visuotactile_system},
we retain the wrist-view images while converting the egocentric view to the robot domain.
SAM~2~\cite{ravi2024sam2} segments the human hands and
wearable hardware, and ProPainter~\cite{zhou2023propainter}
inpaints the masked regions of the egocentric video.
Our tactile-conditioned retargeting method
(Sec.~\ref{sec:tactile_retargeting}) maps the camera-frame
fingertip trajectories and tactile contact observations to
robot hand-base poses and hand joint configurations.
We then render the robot meshes onto the inpainted
egocentric frames using calibrated camera intrinsics
and depth-aware occlusion handling.

\subsubsection{Camera-anchored Action Construction}
\label{sec:camera_anchored_action}

Following UMI~\cite{chi2024umi}, we represent arm actions
as end-effector poses relative to the pose at the start
of each action chunk. Hand actions use absolute joint
targets obtained from retargeting.

We estimate the motion of the head-mounted camera using
an AprilTag fixed in the workspace, separate from the
wrist-mounted fiducials used for hand tracking.
For a chunk anchored at observation time $\tau$, we
transform each retargeted end-effector pose from the camera
frame $C_t$ at time $t$ to the anchor frame $C_\tau$:
\begin{equation}
{}^{C_\tau}\widetilde{\mathbf{T}}_{E_b,t}
=
{}^{C_\tau}\mathbf{T}_{A}
\left(
{}^{C_t}\mathbf{T}_{A}
\right)^{-1}
{}^{C_t}\mathbf{T}_{E_b,t},
\label{eq:camera_anchored_ee}
\end{equation}
where $A$ denotes the stationary scene tag and
$b\in\{L,R\}$ indexes the two arms. These aligned poses
are then used to compute the relative arm actions.

\begin{figure}[t]
    \centering
    \setlength{\abovecaptionskip}{3pt}
\includegraphics[width=0.89\columnwidth]{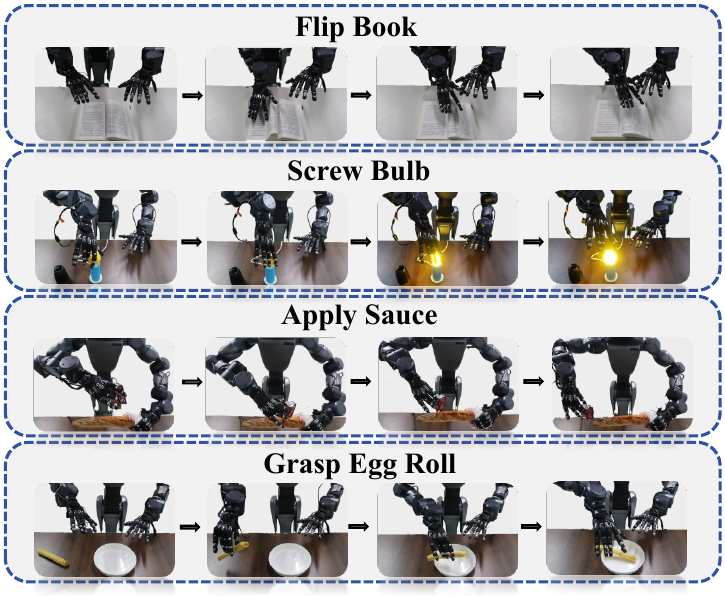}

    \caption{Four contact-rich tasks evaluated in our experiments.}
    \vspace{-20pt}
    \label{fig:task_descriptions}
\end{figure}

\section{Experiments}

\textbf{Overview}: We evaluate TouchBridge through data analyses and real-robot experiments to address four questions:

\begin{enumerate}
    \item How closely do our human-derived robot demonstrations resemble real-robot demonstrations in visual observations and actions?

    \item Can policies fine-tuned solely on our human-derived
demonstrations achieve task performance comparable to that of policies fine-tuned on an equal number of teleoperated
robot demonstrations?

    \item How effective is Tac-Retarget compared with existing simulation-free retargeting methods, and how do its tactile components contribute to its performance?
    
     \item How does policy pretraining affect learning from our human-derived robot demonstrations?

\end{enumerate}

\subsection{Experimental Setup}

\subsubsection{Robot Platform and Policy Execution}
All real-robot evaluations are conducted on a Moz1 bimanual robot equipped with two Wuji Hands. The visual and tactile sensing configuration follows the deployment interface described in Sec.~\ref{sec:robot_interface}. All policies use the same observation acquisition procedures,
low-level controllers, and action-execution rate.

\begin{figure*}[!t]
    \centering

      \includegraphics[width=0.9\textwidth]{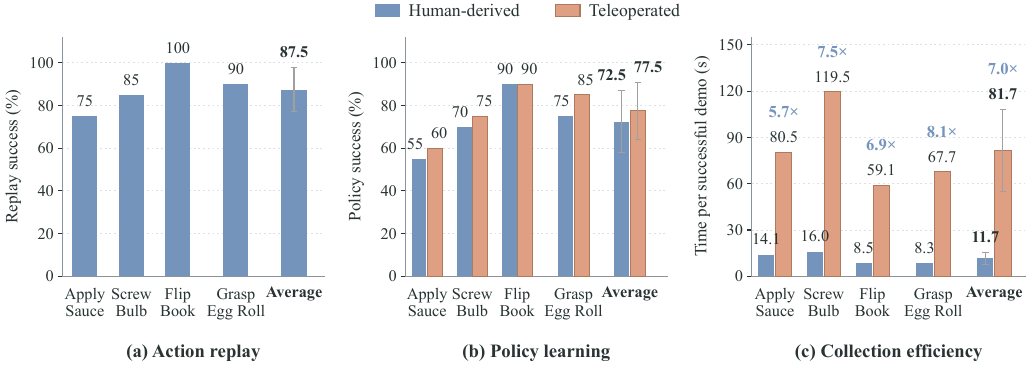}
    \caption{
    (a) Action replay success rates across four tasks,
    with 20 converted human demonstrations replayed per task
    under matched initial scene conditions.
    (b) Real-robot success rates of GR00T N1.7~\cite{gr00tn1_2025} policies
    fine-tuned on 60 demonstrations per task from each data source.
    (c) Average collection time per successful demonstration,
    including failed attempts and scene resets.
    Multipliers indicate collection speedups. Error bars show standard deviation over the four tasks.
    }
    \vspace{-20pt}
    \label{fig:data_source_success}
\end{figure*}

\subsubsection{Task Descriptions}
\label{sec:task_descriptions}

We evaluate all methods on four contact-rich tasks: Flip Book, Screw Bulb, Apply Sauce, and Grasp Egg Roll. These tasks involve diverse contact interactions, from precise fingertip manipulation to force-sensitive object handling. Representative rollouts are shown in Fig.~\ref{fig:task_descriptions}.

\textbf{Flip Book.}
The robot turns a single page of a book using the right index finger.
The motion requires sustained fingertip contact throughout
the page turn while avoiding displacement of the book.

\textbf{Screw Bulb.}
The robot screws a light bulb into a socket until the bulb lights up. 
The task requires stable multi-finger contact and accurate rotational motion.

\textbf{Apply Sauce.}
The robot holds a baguette with its left hand and squeezes ketchup onto
it from left to right with its right hand. The task requires precise
fingertip force control to produce the desired amount of ketchup.

\textbf{Grasp Egg Roll.}
The robot picks up a fragile egg roll with the right hand and places it
onto a plate. The robot must apply sufficient grasp force to prevent slippage without causing visible deformation or breakage.

\subsubsection{Data Collection}
\label{sec:data_collection}

We construct separate human-derived and teleoperated robot datasets for
the four evaluation tasks. The human demonstrations are converted into robot
training sequences using the pipeline in
Sec.~\ref{sec:human_to_robot}. Tactile measurements are used only for retargeting,
not as policy inputs.

For each task, we collect $60$ human demonstrations and $60$ teleoperated robot demonstrations. The teleoperated demonstrations serve as a reference for data-quality analysis and are used to train separate policies for the data-source comparison. They are not mixed into the training data of our method or the human-data baselines. Within each human demonstration, the operator is instructed to minimize head motion to reduce
frame-to-frame viewpoint changes. Across demonstrations, we vary the
camera position and viewing angle to increase the diversity of
camera-object configurations.

\subsubsection{Evaluation Protocol}
\label{sec:evaluation_protocol}

\textbf{Policy Success Rate.}
We evaluate each policy over 20 real-world trials per task.
Success rate is the percentage of trials satisfying the
task-specific success criterion.
\textbf{Replay Success Rate.}
We replay 20 human-derived converted demonstrations per task and report
the percentage of replays satisfying the same task-specific
success criteria used for policy evaluation.
\textbf{Data Collection Time.}
For each task and collection method, we report the average
collection time per successful demonstration by dividing the total elapsed collection time by the 60 successful demonstrations.

\subsection{Evaluation of Human-Derived Robot Demonstrations}
\label{sec:data_quality}

We assess our human-derived robot demonstrations from two
perspectives: visual similarity to real-robot observations and
the ability of the converted actions to reproduce successful
task execution.

\subsubsection{Visual Similarity}

\begin{figure}[t]
    \centering
    \includegraphics[width=\columnwidth]{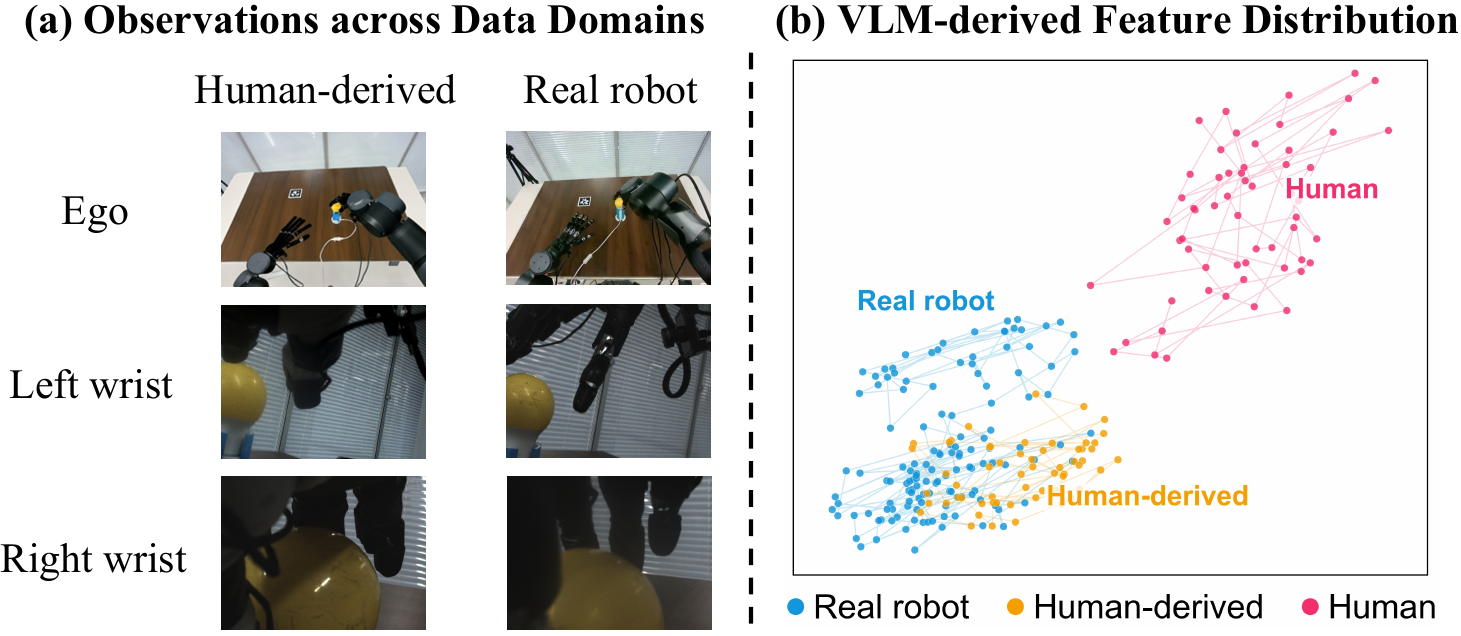}
    \vspace{-10pt}
    \caption{Observation comparison across data domains.
    (a) Multi-view observations from real-robot and human-derived data.
    (b) VLM feature distributions before the action expert.}
    \label{fig:feature_distribution}
    \vspace{-20pt}
\end{figure}

We feed each multi-view observation, together with a shared task
instruction, into the vision-language model (VLM) backbone of GR00T N1.7~\cite{gr00tn1_2025}. We then extract
the output of the final vision--language fusion layer before the
action expert as the observation representation.
As shown in Fig.~\ref{fig:feature_distribution}(b), the projected
features of human-derived robot observations substantially overlap
with those of real-robot observations, whereas raw human
observations exhibit greater separation from the real-robot data.
This pattern suggests that the transformation reduces the visual
gap. Fig.~\ref{fig:feature_distribution}(a) provides a qualitative
comparison of human-derived and real-robot observations
in egocentric and wrist views.

\subsubsection{Action Replay}
We replay the actions from our human-derived robot
demonstrations on the robot, keeping the initial scene
configuration matched to that used during demonstration
collection. Following the replay evaluation protocol in
Sec.~\ref{sec:evaluation_protocol}, the converted actions
achieve an average replay success rate of 87.5\% across the
four tasks (Fig.~\ref{fig:data_source_success}(a)).

Together, these results suggest that our pipeline reduces
the visual gap with real-robot data while retaining
effective actions for task execution.

\begin{figure}[t]
    \centering
\includegraphics[width=0.96\columnwidth]{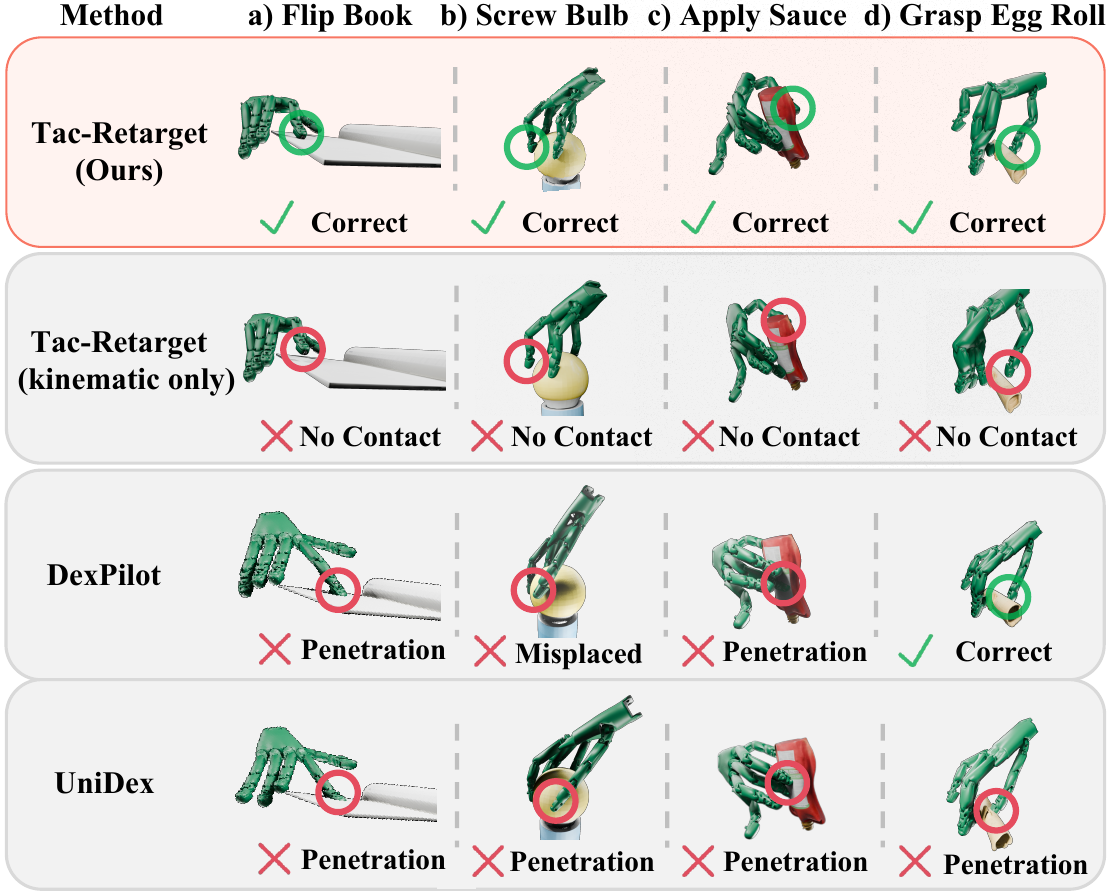}
    \vspace{-3pt}
    \caption{Qualitative comparison of Tac-Retarget, its
    kinematic-only variant, and retargeting baselines.}
    \label{fig:retarget_vis}
        \vspace{-5pt}
\end{figure}

\begin{table}[!t]
    \centering

    \caption{Contact error of retargeting methods and tactile ablations.}
    \label{tab:contact_error}
    \begin{tabular*}{0.95\linewidth}
        {@{\extracolsep{\fill}}lc@{}}
        \toprule
        \textbf{Method} & \textbf{Contact error (mm)} \\
        \midrule
        Tac-Retarget (Ours) & \textbf{4.667} \\
        DexPilot & 6.354 \\
        UniDex & 7.715 \\
        \midrule
        Tac-Retarget (w/o contact graph) & 6.659 \\
        Tac-Retarget (w/o contact alignment) & 7.071 \\
        Tac-Retarget (kinematic only) & 7.540 \\
        \bottomrule
    \end{tabular*}
    \vspace{-15pt}
\end{table}

\subsection{Policy Learning from Human-Derived vs. Robot Demonstrations}
\label{sec:data_source_comparison}

We evaluate whether policies fine-tuned on our human-derived demonstrations perform comparably to those fine-tuned on
teleoperated robot demonstrations. For each task, we fine-tune GR00T N1.7~\cite{gr00tn1_2025} on either 60 human-derived
demonstrations or 60 teleoperated robot demonstrations,
using identical training settings.

As shown in Fig.~\ref{fig:data_source_success}(b), policies trained
on human-derived data achieve an average success rate of 72.5\%,
compared with 77.5\% for those trained on teleoperated robot data.
They achieve the same success rate on Flip Book and remain
within ten percentage points on each of the other three tasks.
These results show that policies fine-tuned solely on our
human-derived demonstrations achieve task success rates
close to those of policies fine-tuned on an equal number
of teleoperated robot demonstrations.

Human demonstration collection is also substantially faster
than robot teleoperation. As shown in
Fig.~\ref{fig:data_source_success}(c), collecting a successful
demonstration takes 11.7 seconds on average with \textbf{TouchBridge-Cap},
compared with 81.7 seconds through teleoperation, corresponding
to a 7.0$\times$ collection speedup. Thus, faster demonstration collection and comparable policy success rates suggest TouchBridge’s potential for scaling robot learning from human demonstrations.

\subsection{Evaluation of Tac-Retarget}
\label{sec:retarget_evaluation}

We evaluate Tac-Retarget, its tactile ablations and baselines through contact preservation
during trajectory replay in simulation and downstream policy
performance on the robot.

\subsubsection{Contact Preservation in Simulation}
We use Grasp Egg Roll for this analysis because it provides a more tractable setting for reliable contact simulation. We load the accurately reconstructed
object mesh into Isaac Sim~\cite{isaacsim} for evaluation.

We apply each retargeting method to the same 20 human demonstrations, replay the resulting
trajectories in the simulation environment and report the contact error of each method in Tab.~\ref{tab:contact_error}. Contact error measures the Euclidean distance between the
response-weighted human tactile contact center and the simulated
contact center in corresponding fingertip-local frames.

As shown in Tab.~\ref{tab:contact_error}, Tac-Retarget achieves
the lowest mean contact error of 4.667~mm, outperforming
DexPilot~\cite{handa2020dexpilot} and UniDex~\cite{zhang2026unidex}. Removing either contact alignment or
the contact graph increases the error, with the kinematic-only
variant performing worst among the ablations. These results show that both components improve contact-location preservation. Fig.~\ref{fig:retarget_vis} illustrates common retargeting
failure modes across four tasks, including missing contacts,
misplaced contacts, and object penetration.
Tac-Retarget preserves the intended fingertip contacts
without obvious object penetration in the examples shown.

\subsubsection{Downstream Policy Performance}

\begin{figure}[t]
    \centering
    \includegraphics[width=\columnwidth]{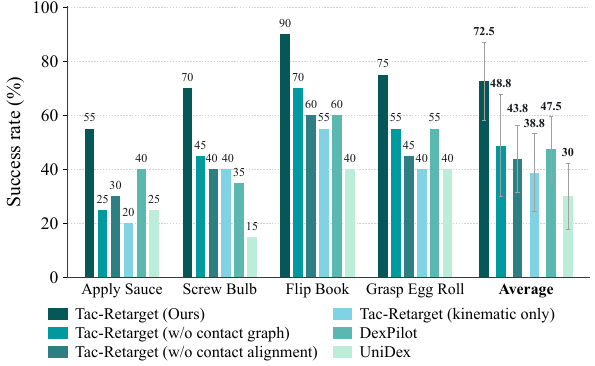}
          \vspace{-15pt}
    \caption{Real-robot policy success rates comparing Tac-Retarget,
    its tactile ablations and existing retargeting methods. Error bars show standard deviation over the four tasks.}
    \label{fig:policy_success_rate}
        \vspace{-20pt}
\end{figure}

For each retargeting method, we convert the same 60 human
demonstrations per task and fine-tune GR00T N1.7~\cite{gr00tn1_2025}
on the resulting data under identical training settings.
Evaluation follows Sec.~\ref{sec:evaluation_protocol}.
As shown in Fig.~\ref{fig:policy_success_rate}, Tac-Retarget
performs best on all four tasks, averaging 72.5\% success
and exceeding DexPilot~\cite{handa2020dexpilot} and
UniDex~\cite{zhang2026unidex} by 25.0 and 42.5 percentage
points, respectively. Removing the contact graph or contact alignment reduces
average success, with the largest drop when both are removed, supporting the contribution
of both tactile components to policy learning.

The simulation and real-robot policy results
show that Tac-Retarget improves contact preservation and
supports more successful policy learning for contact-rich manipulation.

\subsection{Effect of Policy Pretraining}
\label{sec:policy_pretraining}

\begin{figure}[t]
    \centering
    \includegraphics[width=\columnwidth]{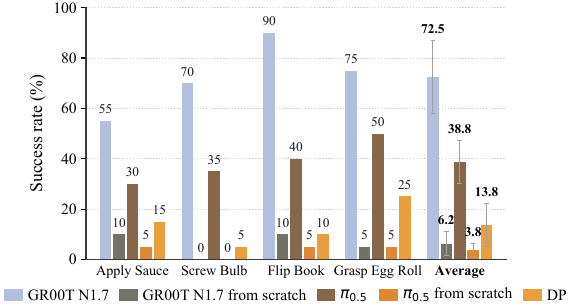}
        \vspace{-15pt}
    \caption{Real-robot policy success rates under different
    policy and pretraining settings. Error bars show standard deviation over the four tasks.}
        \vspace{-20pt}
    \label{fig:policy_pretraining}
\end{figure}

We examine how policy pretraining affects learning from our
human-derived robot demonstrations using the same 60 converted
demonstrations per task. As shown in
Fig.~\ref{fig:policy_pretraining}, pretrained GR00T N1.7~\cite{gr00tn1_2025} and
$\pi_{0.5}$~\cite{black2025pi05} achieve average success rates
of 72.5\% and 38.8\%, respectively, compared with 6.2\% and
3.8\% for their respective from-scratch variants.
Diffusion Policy~\cite{chi2023diffusionpolicy}, included
as an additional baseline, achieves 13.8\%.
All policies are evaluated following
Sec.~\ref{sec:evaluation_protocol}, without task-specific
robot demonstrations for training.
These results demonstrate the benefit of pretraining when
learning from limited human-derived demonstrations, with
pretrained GR00T N1.7~\cite{gr00tn1_2025} achieving the best performance
among the evaluated policies.

\section{Conclusion}

We presented TouchBridge, a hardware--software co-designed
framework for learning contact-rich dexterous manipulation
without task-specific robot demonstrations. It combines
TouchBridge-Cap for efficient human demonstration collection
with Tac-Retarget for contact-preserving motion transfer,
producing robot-domain data without object meshes,
object-pose estimates, or simulation-based refinement.
With faster demonstration collection and policy success
rates close to those obtained from robot-collected data,
we believe TouchBridge can serve as a practical framework
for the community to scale robot learning from human
demonstrations. Our current study uses a modest demonstration
set; extending the framework to larger and more diverse
human datasets remains an important direction for future work.


\section*{APPENDIX}

\subsection{Multimodal Data Collection and Temporal Alignment}
\label{app:data_acquisition_alignment}

\subsubsection{Asynchronous Multimodal Collection}

Human demonstrations are recorded using independent acquisition
threads for egocentric RGB-D observations, bilateral wrist-view
images, articulated hand poses, and bilateral fingertip tactile
measurements. All samples are timestamped using a shared host clock.
Table~\ref{tab:acquisition_rates} summarizes the collecting
sampling rates of the multimodal streams.
\begin{table}[t]
    \centering
    \caption{Collecting rates of the human demonstration streams.}
    \label{tab:acquisition_rates}
    \footnotesize
    \setlength{\tabcolsep}{3pt}
    \renewcommand{\arraystretch}{1.08}
    \begin{tabularx}{\columnwidth}{
        @{}>{\raggedright\arraybackslash}X
        >{\centering\arraybackslash}p{0.28\columnwidth}@{}
    }
        \toprule
        Data stream & Collecting rate \\
        \midrule
        Egocentric RGB
            & 30~Hz \\
        Left/right wrist-view RGB
            & 30~Hz \\
        Bilateral Manus hand poses
            & 130~Hz \\
        Fingertip tactile stream
            & 90~Hz \\
        \bottomrule
    \end{tabularx}
\end{table}

\begin{figure*}[t]
    \centering
    \includegraphics[width=0.95\textwidth]
    {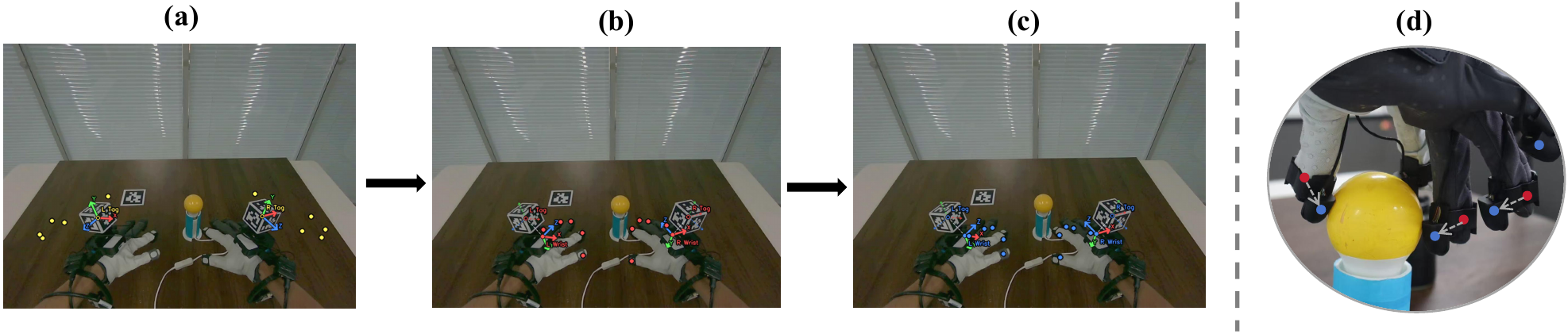}
    \caption{\textbf{Reconstruction of camera-frame tactile
    fingertip poses.}
    (a) Wrist-mounted AprilTag fiducial frames detected in the
    egocentric image.
    (b) Manus fingertip positions (red) reconstructed using the
    calibrated wrist-to-cube transforms.
    (c) Tactile-sensor origins (blue) obtained by applying the
    fixed sensor-to-fingertip mounting transform.
    (d) Close-up of the fixed spatial offsets between the Manus
    fingertips (red) and tactile-sensor origins (blue).}
    \label{fig:fingertip_pose_reconstruction}
\end{figure*}

\subsubsection{Camera-Frame Fingertip Reconstruction}
\label{app:fingertip_pose_reconstruction}

As shown in Fig.~\ref{fig:fingertip_pose_reconstruction},
we first estimate the AprilTag fiducial-cube pose
${}^{C}\mathbf{T}_{Q}$ in the egocentric camera frame.
Given the Manus fingertip pose
${}^{W}\mathbf{T}_{F_i}$ relative to the wrist, its
camera-frame pose is reconstructed as
\begin{equation}
    {}^{C}\mathbf{T}_{F_i}
    =
    {}^{C}\mathbf{T}_{Q}
    {}^{Q}\mathbf{T}_{W}
    {}^{W}\mathbf{T}_{F_i},
    \label{eq:fingertip_reconstruction}
\end{equation}
where $C$, $Q$, $W$, and $F_i$ denote the camera,
fiducial-cube, wrist, and $i$-th fingertip frames,
respectively.

Because the tactile sensor is mounted with a fixed offset
from the Manus fingertip frame, we further compute its pose
as
\begin{equation}
    {}^{C}\mathbf{T}_{S_i}
    =
    {}^{C}\mathbf{T}_{F_i}
    {}^{F_i}\mathbf{T}_{S_i},
    \label{eq:sensor_reconstruction}
\end{equation}
where $S_i$ denotes the corresponding tactile-sensor frame.
The resulting sensor origin and orientation are used as the
human fingertip target in Tac-Retarget, such that the
kinematic target is expressed in the same coordinate frame
as the tactile taxel layout.

For the hardware configuration shown in
Fig.~\ref{fig:fingertip_pose_reconstruction}, the calibrated
wrist-to-cube transform is
\begin{equation}
{}^{Q}\mathbf{T}_{W}
=
\begin{bmatrix}
 1 &  0 &  0 & 0.000 \\
 0 & -1 &  0 & -0.065 \\
 0 &  0 & -1 & -0.015 \\
 0 &  0 &  0 & 1
\end{bmatrix}.
\label{eq:wrist_to_cube}
\end{equation}

The sensor-to-fingertip mounting transform, shared by all
fingertips, is
\begin{equation}
{}^{F_i}\mathbf{T}_{S_i}
=
\begin{bmatrix}
-1 & 0         & 0        & 0       \\
 0 & -0.275637 & 0.961262 & 0.02477 \\
 0 &  0.961262 & 0.275637 & 0.00186 \\
 0 & 0         & 0        & 1
\end{bmatrix}.
\label{eq:sensor_to_fingertip}
\end{equation}
All translation entries are expressed in meters.

\subsection{Implementation Details of Tac-Retarget}
\label{app:tac_retarget_details}

This section supplements Sec.~\ref{sec:tactile_retargeting}
with tactile preprocessing, contact-dependent weighting,
graph construction, and compliance compensation.

\subsubsection{Tactile Preprocessing and Contact Activation}

We align the tactile measurements with the visual timestamps
by linearly interpolating the three-axis taxel readings.
For taxel $m$ on fingertip $i$, its response magnitude is
$s_{im}=\|\mathbf{f}_{im}\|_2$, and the total fingertip
response is $F_i=\sum_m s_{im}$.
The active-taxel set is
$\mathcal{A}_i=\{m:s_{im}>\tau_s\}$.
A contact label $c_i\in\{0,1\}$ is set to one when
$F_i>\tau_F$ and $|\mathcal{A}_i|\geq n_{\min}$.
The thresholds are expressed in the input tactile-response
scale.

The contact center in Eq.~\eqref{eq:patch_center} uses
responses from all taxels, whereas the taxel-cloud term
uses only $\mathcal{A}_i$.
For each active taxel, its weight is
\begin{equation}
    \omega_{im}
    =
    \frac{1}{|\mathcal{A}_i|}
    \frac{s_{im}}
    {\max\!\left(\max_{k\in\mathcal{A}_i}s_{ik},\epsilon\right)}.
    \label{eq:appendix_taxel_weight}
\end{equation}
This weighting emphasizes stronger responses while
averaging the contribution over the active taxels.
We use the same calibrated local taxel layout for the human and robot hands. The contact center, represented in this common local sensor frame, is transformed into the camera frame using the corresponding human or robot sensor pose.

\subsubsection{Contact-Conditioned Weighting and Graph Construction}

Let $c_i\in\{0,1\}$ indicate whether the $i$-th
fingertip is in contact. We use the contact-conditioned
pose-tracking weight for fingertips:
\begin{equation}
    w_i
    =
    \lambda_p
    (\beta_0+\beta_c c_i)\eta^{c_i},
    \label{eq:appendix_contact_weights}
\end{equation}
where $\eta$ reduces pose tracking during detected contact.
The tactile patch, taxel-cloud, and compliance objectives
are activated upon contact and weighted by their respective
global coefficients.

The rotation coefficient is configured independently as
$\lambda_R$, corresponding to
$\lambda_{\mathrm{rot}}=\lambda_R/\lambda_p$
in Eq.~\eqref{eq:fingertip_pose_alignment}.
With $\beta_0=1$, $\beta_c=2$, and $\eta=0.05$, the
pose-tracking weight during contact is $0.15$ times its
non-contact value. This reduction allows the tactile
objectives to influence the solution while preserving
fingertip motion tracking.

The response $\phi_u$ in
Eq.~\eqref{eq:tactile_graph_weight} is the associated
fingertip total response for fingertip and contact-center
nodes.
Human-derived edge weights are reused for the robot graph.
In addition to Laplacian matching, the implementation
penalizes discrepancies in Euclidean distances between
all pairs of contacting fingertips, with an additional
coefficient for cross-hand pairs.
These auxiliary terms preserve contact spacing within
and across hands.

\subsubsection{Response-Dependent Compliance Compensation}

Contact-patch alignment preserves the demonstrated
contact geometry, but does not explicitly represent local
deformation during contact.
We therefore include a response-dependent geometric
compensation term that shifts the target contact center.
For an active contact, the offset magnitude is
\begin{equation}
    d_i
    =
    \min\!\left(
        d_{\max},
        \frac{\max(0,F_i-F_{\min})}{k}
    \right),
    \label{eq:appendix_compliance_depth}
\end{equation}
where $k$ controls the response-to-displacement conversion
and $d_{\max}$ bounds the target offset.

Using the human sensor origin $\mathbf{p}_i^H$ and contact center
$\boldsymbol{\mu}_i^H$, we define
\begin{equation}
\begin{aligned}
    \mathbf{u}_i
    &=
    \frac{\boldsymbol{\mu}_i^H-\mathbf{p}_i^H}
    {\|\boldsymbol{\mu}_i^H-\mathbf{p}_i^H\|_2},\\
    \boldsymbol{\mu}_i^{H,\mathrm{comp}}
    &=
    \boldsymbol{\mu}_i^H+d_i\mathbf{u}_i ,
\end{aligned}
\label{eq:appendix_compliance_target}
\end{equation}
where $\mathbf{u}_i$ points from the sensor toward the contacted object,
so that the compliance target lies slightly beyond the measured contact
surface. Intuitively, once contact is established, the fingertip is
encouraged to press slightly deeper, as if the compliant fingertip pad
were compressed under load. The displacement increases with the tactile
response up to a prescribed maximum depth, encouraging a firmer and more
stable grasp rather than merely touching the object surface.

The implementation augments the objective in
Eq.~\eqref{eq:tactile_retarget} with
\begin{equation}
    \lambda_{\mathrm{comp}}\mathcal{L}_{\mathrm{comp}}
    =
    \lambda_{\mathrm{comp}}
    \sum_{i:c_i=1}
    \left\|
        \boldsymbol{\mu}_i^R
        -
        \boldsymbol{\mu}_i^{H,\mathrm{comp}}
    \right\|_2^2 .
    \label{eq:appendix_compliance_loss}
\end{equation}

\subsubsection{Hyperparameters}
\label{app:tac_retarget_optimization}

Table~\ref{tab:tac_retarget_hyperparameters} summarizes the
selected parameters that directly correspond to the
objectives described above.
The table reports separate absolute weights for patch-center
and active-taxel alignment. In the factorized notation of
Eq.~\eqref{eq:tactile_contact_alignment}, these correspond
to $\lambda_C=\lambda_{\mathrm{patch}}$ and
$\lambda_x=\lambda_{\mathrm{taxel}}/
\lambda_{\mathrm{patch}}$.

\begin{table*}[t]
    \centering
    \caption{Selected hyperparameters for Tac-Retarget.}
    \label{tab:tac_retarget_hyperparameters}
    \footnotesize
    \setlength{\tabcolsep}{6pt}
    \renewcommand{\arraystretch}{1.08}
    \vspace{+6pt}
    \begin{tabularx}{\textwidth}{
        @{}>{\raggedright\arraybackslash}p{0.16\textwidth}
        >{\centering\arraybackslash}p{0.13\textwidth}
        >{\raggedright\arraybackslash}X@{}
    }
        \toprule
        \textbf{Parameter} & \textbf{Value} & \textbf{Description} \\
        \midrule
        $\tau_F$
            & $0.02$
            & Contact threshold on the total fingertip response. \\
        $\tau_s$
            & $0.01$
            & Response threshold for selecting active taxels. \\
        $\lambda_p$
            & $100$
            & Camera-frame fingertip-position tracking weight. \\
        $\lambda_R$
            & $1$
            & Fingertip-orientation tracking weight. \\
        $(\beta_0,\beta_c)$
            & $(1,\,2)$
            & Baseline tracking multiplier and contact-dependent increment. \\
        $\eta$
            & $0.05$
            & Pose-tracking scale applied during detected contact. \\
        $\lambda_{\mathrm{patch}}$
            & $20$
            & Contact-patch center alignment weight. \\
        $\lambda_{\mathrm{taxel}}$
            & $4$
            & Active-taxel cloud alignment weight. \\
        $\lambda_{\mathrm{comp}}$
            & $16$
            & Response-dependent compliance-target weight. \\
        $k$
            & $20$
            & Response-to-displacement conversion factor. \\
        $d_{\max}$
            & $0.01$~m
            & Maximum inward compliance displacement. \\
        $\lambda_G$
            & $1$
            & Tactile contact-graph Laplacian weight. \\
        $\kappa$
            & $20$~m$^{-1}$
            & Distance-decay factor for graph-edge weights. \\
        $\gamma$
            & $1$
            & Tactile-response modulation of graph-edge weights. \\
        $(\lambda_{\mathrm{cross}},\lambda_{\mathrm{pair}})$
            & $(0.4,\,0.25)$
            & Cross-hand and pairwise contact-distance weights. \\
        $\lambda_{q,\mathrm{smooth}}$
            & $0.45$
            & First-order temporal smoothness weight for joint angles. \\
        $(\lambda_{b,p},\lambda_{b,R})$
            & $(10,\,5)$
            & First-order smoothness weights for base translation and rotation. \\
        $\lambda_{\mathrm{acc}}$
            & $0.08$
            & Second-order smoothness weight for bases and joint angles. \\
        \bottomrule
    \end{tabularx}
\end{table*}
\section*{ACKNOWLEDGMENT}

This work is supported by Tsinghua University - Keystone Electrical (Zhejiang) Co.,Ltd Joint Research Center for Embodied Multimodal Artificial Intelligence (JCEMAI).

\addtolength{\textheight}{-1cm}   


\begin{thebibliography}{99}

\bibitem{zhang2026unidex}
G. Zhang, Q. Xu, H. Zhang, J. Ma, L. He, Y. Bao, Z. Ping, Z. Yuan,
C. Lu, C. Yuan, \emph{et al.},
``UniDex: A robot foundation suite for universal dexterous hand control
from egocentric human videos,''
in \emph{Proc. IEEE/CVF Conf. Comput. Vis. Pattern Recognit. (CVPR)},
2026, pp. 1841--1852.

\bibitem{zheng2026egoscale}
R. Zheng \emph{et al.},
``EgoScale: Scaling dexterous manipulation with diverse egocentric
human data,''
\emph{arXiv preprint arXiv:2602.16710}, 2026.

\bibitem{wang2026humanego}
Z. Wang, B. He, K. Yu, S. Lee, R. Gao, F. Huang, and Y. Aloimonos,
``HumanEgo: Zero-shot robot learning from minutes of human egocentric
videos,''
\emph{arXiv preprint arXiv:2605.24934}, 2026.

\bibitem{guzey2026dexterity}
I. Guzey, H. Qi, J. Urain, C. Wang, J. Yin, K. Bodduluri, M. Lambeta,
L. Pinto, A. Rai, J. Malik, T. Wu, A. Sharma, and H. Bharadhwaj,
``Dexterity from smart lenses: Multi-fingered robot manipulation with
in-the-wild human demonstrations,''
in \emph{Proc. IEEE Int. Conf. Robot. Autom. (ICRA)}, 2026.

\bibitem{handa2020dexpilot}
A. Handa, K. Van Wyk, W. Yang, J. Liang, Y.-W. Chao, Q. Wan,
S. Birchfield, N. D. Ratliff, and D. Fox,
``DexPilot: Vision-based teleoperation of dexterous robotic hand-arm
system,''
in \emph{Proc. IEEE Int. Conf. Robot. Autom. (ICRA)}, 2020,
pp. 9164--9170.

\bibitem{qin2023anyteleop}
Y. Qin, W. Yang, B. Huang, K. Van Wyk, H. Su, X. Wang,
Y.-W. Chao, and D. Fox,
``AnyTeleop: A general vision-based dexterous robot arm-hand
teleoperation system,''
in \emph{Proc. Robot.: Sci. Syst. (RSS)}, 2023,
doi: 10.15607/RSS.2023.XIX.015.

\bibitem{huang2025fungrasp}
L. Huang, H. Zhang, Z. Wu, S. Christen, and J. Song,
``FunGrasp: Functional grasping for diverse dexterous hands,''
\emph{IEEE Robot. Autom. Lett.}, vol. 10, no. 6,
pp. 6175--6182, Jun. 2025,
doi: 10.1109/LRA.2025.3561573.

\bibitem{lin2025dexflow}
X. Lin, K. Yao, L. Xu, X. Wang, X. Li, Y. Wang, and M. Li,
``DexFlow: A unified approach for dexterous hand pose retargeting
and interaction,''
\emph{arXiv preprint arXiv:2505.01083}, 2025.

\bibitem{wu2026toporetarget}
J. Wu, S. Yao, G. He, X. Liu, Z. Zeng, X. Jiang, H. Yang,
W. Zhang, and H. Zhao,
``TopoRetarget: Interaction-preserving retargeting for dexterous
manipulation,''
\emph{arXiv preprint arXiv:2606.16272}, 2026.

\bibitem{fan2026robopaint}
J. Fan, Z. Zhao, Y. Zhang, C. Chen, P. Wang, H. Zhang, and Z. Cheng,
``RoboPaint: From human demonstration to any robot and any view,''
\emph{arXiv preprint arXiv:2602.05325}, 2026.

\bibitem{li2025maniptrans}
K. Li, P. Li, T. Liu, Y. Li, and S. Huang,
``ManipTrans: Efficient dexterous bimanual manipulation transfer via
residual learning,''
in \emph{Proc. IEEE/CVF Conf. Comput. Vis. Pattern Recognit. (CVPR)},
2025, pp. 6991--7003.

\bibitem{kedia2026simtoolreal}
K. Kedia, T. G. W. Lum, J. Bohg, and C. K. Liu,
``SimToolReal: An object-centric policy for zero-shot dexterous tool
manipulation,''
\emph{arXiv preprint arXiv:2602.16863}, 2026.


\bibitem{kareer2025egomimic}
S. Kareer, D. Patel, R. Punamiya, P. Mathur, S. Cheng, C. Wang,
J. Hoffman, and D. Xu,
``EgoMimic: Scaling imitation learning via egocentric video,''
in \emph{Proc. IEEE Int. Conf. Robot. Autom. (ICRA)}, 2025,
pp. 13226--13233.

\bibitem{yuan2026motiontrans}
C. Yuan, R. Zhou, M. Liu, Y. Hu, S. Wang, L. Yi, C. Wen, S. Zhang,
and Y. Gao,
``MotionTrans: Human VR data enable motion-level learning for robotic
manipulation policies,''
in \emph{Proc. IEEE Int. Conf. Robot. Autom. (ICRA)}, 2026.

\bibitem{wang2025boostergym}
Y. Wang, P. Chen, X. Han, F. Wu, and M. Zhao,
``Booster Gym: An end-to-end reinforcement learning framework for humanoid
robot locomotion,''
\emph{arXiv preprint arXiv:2506.15132}, 2025.

\bibitem{wang2026soccer}
Y. Wang, C. Luo, P. Chen, J. Liu, W. Sun, T. Guo, K. Yang, B. Hu, Y. Zhang,
and M. Zhao,
``Learning vision-driven reactive soccer skills for humanoid robots,''
\emph{Sci. Robot.}, vol. 11, 2026.

\bibitem{chen2026dexx}
R. Chen, F. Ruan, L. Cao, Z. Wang, B. Xu, S. Tong, J. Liu, M. Pei,
C. Zhang, W. Xing, K. Zhang, and M. Xu,
``Dex-X: Learning visual-tactile dexterous manipulation from human videos
with simulated interaction,''
\emph{arXiv preprint arXiv:2609.07747}, 2026.

\bibitem{si2024difftactile}
Z. Si, G. Zhang, Q. Ben, B. Romero, Z. Xian, C. Liu, and C. Gan,
``DiffTactile: A physics-based differentiable tactile simulator for
contact-rich robotic manipulation,''
in \emph{Proc. Int. Conf. Learn. Represent. (ICLR)}, 2024.

\bibitem{xu2025dexumi}
M. Xu, H. Zhang, Y. Hou, Z. Xu, L. Fan, M. Veloso, and S. Song,
``DexUMI: Using human hand as the universal manipulation interface for
dexterous manipulation,''
in \emph{Proc. Conf. Robot Learn. (CoRL)}, 2025,
pp. 437--459.

\bibitem{huang2026icl}
H. Huang \emph{et al.},
``In-Context Learning for Robots: Methods and Applications,''
\emph{arXiv preprint arXiv:2609.36012}, 2026.

\bibitem{jiang2024cimpc}
Y. Jiang, M. Yu, X. Zhu, M. Tomizuka, and X. Li,
``Contact-implicit model predictive control for dexterous in-hand manipulation:
A long-horizon and robust approach,''
in \emph{Proc. IEEE/RSJ Int. Conf. Intell. Robots Syst. (IROS)}, 2024.

\bibitem{wang2026univlat}
Z. Wang, S. Liu, S. Zheng, L. Cao, R. Chen, R. Liu, Y. Yang, and M. Xu,
``Uni-VLaT: Whole-body tactile adaptation of VLA policies for humanoid
loco-manipulation,''
\emph{arXiv preprint arXiv:2609.35450}, 2026.

\bibitem{ren2026c2dex}
J. Ren, Z. Jiang, Y. Yang, H. Jia, H. Jiang, B. Li, Y. Yao,
C. Lin, Q. Shen, Z. Bing, X.-X. Long, and X. Cao,
``C2Dex: Contact-consistent reconstruction and retargeting for
dexterous manipulation from monocular video,''
\emph{arXiv preprint arXiv:2608.07045}, 2026.

\bibitem{spider}
C. Pan, C. Wang, H. Qi, J. Urain, Z. Liu, H. Bharadhwaj, A. Sharma, T. Wu,
G. Shi, J. Malik, and F. Hogan,
``SPIDER: Scalable physics-informed dexterous retargeting,''
\emph{arXiv preprint arXiv:2511.09484}, 2025.

\bibitem{dai2026card}
C. Dai, B. He, F. R. Hogan, W. Pan, R. Fu, W.-H. Hsieh, R. Liu, J. Malik,
G. Shi, and A. Sharma,
``CARD: Contact-anchored retargeting for dexterous manipulation,''
\emph{arXiv preprint arXiv:2610.36539}, 2026.

\bibitem{ravi2024sam2}
N. Ravi \emph{et al.},
``SAM 2: Segment Anything in Images and Videos,''
\emph{arXiv preprint arXiv:2408.00714}, 2024.

\bibitem{zhou2023propainter}
S. Zhou, C. Li, K. C. K. Chan, and C. C. Loy,
``ProPainter: Improving propagation and transformer for video
inpainting,''
in \emph{Proc. IEEE/CVF Int. Conf. Comput. Vis. (ICCV)}, 2023,
pp. 10477--10486.

\bibitem{chi2024umi}
C. Chi, Z. Xu, C. Pan, E. Cousineau, B. Burchfiel, S. Feng,
R. Tedrake, and S. Song,
``Universal Manipulation Interface: In-the-wild robot teaching without
in-the-wild robots,''
in \emph{Proc. Robot.: Sci. Syst. (RSS)}, 2024.

\bibitem{gr00tn1_2025}
NVIDIA \emph{et al.},
``GR00T N1: An open foundation model for generalist humanoid robots,''
\emph{arXiv preprint arXiv:2503.14734}, 2025.

\bibitem{isaacsim}
S. Gao, M. Pagnucco, T. Bednarz, and Y. Song,
``NVIDIA Isaac Sim: Enabling scalable, GPU-accelerated simulation
for robotics,''
\emph{arXiv preprint arXiv:2606.03551}, 2026.

\bibitem{black2025pi05}
K. Black \emph{et al.},
``$\pi_{0.5}$: A vision-language-action model with open-world
generalization,''
in \emph{Proc. Conf. Robot Learn. (CoRL)}, 2025,
pp. 17--40.

\bibitem{chi2023diffusionpolicy}
C. Chi, S. Feng, Y. Du, Z. Xu, E. Cousineau, B. Burchfiel,
and S. Song,
``Diffusion Policy: Visuomotor policy learning via action diffusion,''
in \emph{Proc. Robot.: Sci. Syst. (RSS)}, 2023,
doi: 10.15607/RSS.2023.XIX.026.



\end{thebibliography}
\end{document}